\documentclass[letterpaper, 10 pt, journal, twoside]{IEEEtran}

\IEEEoverridecommandlockouts

\usepackage[T1]{fontenc}

\usepackage{amsmath,amssymb,amsfonts}
\usepackage{algorithmic}
\usepackage{algorithm}
\usepackage{graphicx}
\usepackage{textcomp}
\usepackage[table,xcdraw]{xcolor}
\usepackage{comment}
\usepackage[sorting=none, firstinits=true, maxbibnames=99,style=numeric-comp]{biblatex}
\usepackage{booktabs}
\usepackage{svg}
\newcommand{\ssymbol}[1]{^{\@fnsymbol{#1}}}

\usepackage[hidelinks]{hyperref}

\definecolor{black}{rgb}{0, 0, 0}
\definecolor{red}{rgb}{0.9, 0, 0}
\definecolor{green}{rgb}{0, 0.6, 0}
\definecolor{blue}{rgb}{0, 0, 0.9}
\definecolor{grey}{rgb}{0.52, 0.52, 0.51}

\newcommand{\BLACK}[1]{\textcolor{black}{#1}}

\newcommand{\rev}[1]{\BLACK{#1}}

\newtheorem{requirement}{\textbf{Requirement}}
\newtheorem{question}{\textbf{Question}}

\begin{document}

\title{Adaptive Human-Robot Collaborative Painting Combining Preference-Based Optimization and Dynamic Motion Primitives}
%
%
%

\author{C. Cella$^{1}$, M. Ristic$^{1}$, M. Faroni$^{1}$, A. M. Zanchettin$^{1}$, and P. Rocco$^{1}$
\thanks{$^{1}$The authors are with Politecnico di Milano, 20133, Milano, Italy ({\footnotesize e-mail: christian.cella@polimi.it}).}
}

\markboth{IEEE Robotics and Automation Letters. Preprint Version. Accepted March, 2026}
{Cella \MakeLowercase{\textit{et al.}}: Adaptive HRC Painting Combining PBO and DMPs}

\maketitle

\begin{abstract}
This work presents a human-centered collaborative framework that integrates Preference-Based Optimization (PBO) and Dynamic Movement Primitives (DMPs) to optimize robot-assisted tasks such as painting. 
The system allows the operator to perform the process while the robot adapts its behavior in real-time, dynamically adjusting the orientation of the piece in order to match the orientation of the operator's hand. 
The PBO framework leverages the GLISp algorithm to iteratively refine control parameters such as execution time, robot responsiveness, and rotation amplification through human feedback. 
Moreover, DMPs have been modified to enhance the reactive behavior of the robot and its adaptability to ergonomic requirements. 
The method was validated with a heterogeneous group of participants executing \rev{painting tasks}. 
The results show that our strategy effectively reduces operator effort while optimizing process outcomes.
\end{abstract}

\begin{IEEEkeywords}
human-robot collaboration, human factors and human-in-the-loop.
\end{IEEEkeywords}

\IEEEpeerreviewmaketitle

\section{Introduction}
\label{sec:intro}
\IEEEPARstart{T}{he} Industry 5.0 paradigm has advanced industrial robotic systems for collaborative processes, enhancing human performance by combining operator dexterity with robotic repeatability \cite{gulzow2020recent}. 
Among these, a key subset involves \emph{non-contact} assistive tasks, where robots support operators without direct physical interaction. Such frameworks improve \rev{efficiency} and ergonomics leveraging \emph{leader-follower} paradigms, where the robot trajectories are optimized \rev{to} ensure adaptive responsiveness to human intent in industrial environments \cite{van2020adaptive}.

In this context, spray painting is one of the most critical industrial applications. 
In sectors like automotive and furniture manufacturing, the quality of the painted surface is essential, influencing both aesthetics and structural integrity \cite{muzan2012implementation}. 
Although automated solutions exist, human expertise remains indispensable, particularly for irregularly shaped or cumbersome components. 
On the other side, prolonged manual operations increase fatigue and ergonomic strain, reducing efficiency over time \cite{zanchettin2019collaborative}. 
The most critical aspect related to traditional spray painting setups is the lack of metrics accounting for operator preferences and feedback, which limits process flexibility and hinders effective robotic assistance across diverse production scenarios. 
The balancing of productivity and ergonomics should transform the robot into an \emph{assistive} partner, capable of adjusting its actions to support the operator's needs during the process, rather than being a passive tool.

\begin{figure}
\centering
\includegraphics[width=0.43\textwidth]{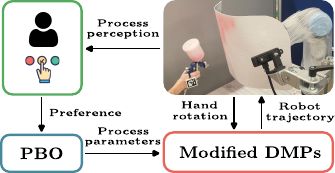}
\caption{
Sketch of the proposed approach applied to collaborative painting.
}
\label{fig:semplified_scheme}
\end{figure}

To address this problem, we propose a \emph{human-centric} robotic assistance framework for non-contact industrial operations. 
Although applicable to various processes, spray painting is our primary focus. 
To this purpose, consider the scenario in Fig. \ref{fig:semplified_scheme}: \rev{the} robot holds \rev{the} workpiece while the operator is painting it.
The robot shall assist the \rev{operator} by adjusting the workpiece pose online, in order to reduce human effort and enhance flexibility. 
Our method is based on a combination of \emph{Preference-Based Optimization} (PBO) and \emph{Dynamic Movement Primitives} (DMPs). 
\rev{While DMPs provide a convenient framework for adaptive trajectory generation, they do not explicitly account for subjective user preferences, which are central in Human-Robot Collaboration (HRC). 
For this purpose, we integrate DMPs-based motion generation with PBO, enabling adjustments of the robot behavior based on qualitative operator feedback.
This approach allows for real-time tuning of robot responsiveness (red box in Fig. \ref{fig:semplified_scheme}), while the GLISp algorithm \cite{bemporad2021global} is employed in between experiments to optimize the set of process parameters from qualitative feedback (green and blue boxes).}
Results on a real-world painting setup show that our approach increases ergonomics and the operator's satisfaction, without jeopardizing the process quality.

\section{Related work and contributions}
\label{sec:state_art}
Assistive robotic tasks in industrial settings require robots to adapt their behavior to human preferences, rather than executing predefined trajectories \cite{kupcsik2018learning}. 
In these cases, ensuring seamless collaboration requires balancing human comfort and process efficiency, a challenge that is widely emphasized in ergonomics-oriented \rev{HRC} strategies, where the goal is to adapt the robot's behavior as a function of the operator's perceived comfort \cite{proia2021control, lorenzini2023ergonomic}. 
In accordance \rev{with} this vision, recent advances integrate \emph{preference-based learning} to personalize the robot actions \cite{castro2021trends}, allowing \rev{them} to adapt their behavior based on \emph{qualitative} human feedback.

Among the industrial applications that could benefit the most from these approaches, robotic painting proposes unique challenges, due to its strict quality requirements and variability in task conditions. 
The adoption of industrial robots in painting improves productivity and surface quality, particularly in the automotive and furniture sectors \cite{muzan2012implementation}. 
Although recent research has introduced collision-free trajectory planning \cite{zbiss2024automatic}, these approaches often overlook human preferences and key aspects in HRC. 
\rev{Indeed, most trajectory generation algorithms focus on geometric optimization while overlooking real-time adaptation to human input and the integration of qualitative feedback, limiting their flexibility and effectiveness in HRC industrial applications \cite{schaldenbrand2024cofrida}.}

A promising solution is \rev{offered by} Preference-Based Optimization (PBO) \cite{bemporad2021global}, which accounts for the operator's preferences with the aim of optimizing the robotic behavior. 
PBO has been successfully applied in HRC scenarios to fine-tune control parameters based on user preferences \cite{maccarini2022preference}. 
Recent works focused on ergonomic optimization, such as fostering operator trust \cite{campagna2024promoting} and integrating ergonomic assessments to enhance comfort and productivity in object-handling tasks \cite{falerni2024framework}. 
From an industrial perspective, PBO has also been employed to optimize velocity planners in \rev{sealing} operations, refining motion control based on \rev{human} preferences \cite{roveda2021pairwise}.

While \emph{preference learning} effectively captures human intent, an additional challenge in HRC applications is represented by the mapping of these preferences into executable motion commands. 
This is where Dynamic Movement Primitives (DMPs) provide \rev{an effective} solution. 
DMPs enable smooth, adaptive trajectory generation, responding to external changes through a dynamical system formulation \cite{ijspeert2002movement}. 
As highlighted in \cite{saveriano2023dynamic}, many extensions have been introduced over the years, ranging from online adaptation mechanisms to formulations that incorporate \emph{coupling terms} for reactive behavior. 
Within this family of improvements, recent works have proposed adaptive coefficients to modulate execution time and end-position behaviors in collaborative handover tasks \cite{iori2023dmp, perovic2023adaptive}. 
The main drawback of these DMPs-based works concerns the type of applications for which they are engineered, where the main interest is to adapt the process to the operator, rather than enhancing the cooperation between the human and the robot based on qualitative feedback coming from the subjective perception of the process.
On the other hand, PBO-based works mainly focus on the estimation of control parameters in physical interaction, \rev{and} they do not \rev{straightforwardly} apply to motion planning for non-contact collaborative tasks. 

In this regard, the integration of DMPs and PBO is an underexplored, yet promising, direction for tasks requiring adaptability and subjective optimization. 
This work aims at filling \rev{this gap} with the following contributions:
\begin{itemize}
    \item  We developed a \emph{human-centric} collaborative painting application that redefines the role of the robot, from a primary executor to an adaptive assistant. 
    Unlike previous approaches, such as \cite{zbiss2022automatic} and \cite{muzan2012implementation}, which primarily focus on autonomous task execution, our pipeline is specifically designed to support the operator during the manipulation of cumbersome components. 
    Additionally, despite both the method presented in \cite{zanchettin2019collaborative} and ours are applicable in the presence of \emph{extended} surfaces, we adapt the robot motion through tunable parameters and preference-driven adjustments, enabling a more flexible and \rev{user-tailored} collaborative painting experience.
    \item We introduced significant modifications to the standard DMPs framework, incorporating rotational amplification and responsiveness adjustments. 
    Taking inspiration from the works dealing with collaborative handovers \cite{iori2023dmp, perovic2023adaptive}, which primarily address spatial and temporal synchronization between the agents, we introduce instead modifications to the DMPs equations that enable the robot to dynamically amplify the operator's hand rotations, while preserving smooth trajectory generation.
\end{itemize}

\section{Problem statement}
\label{sec:problem_statement}
Painting tasks can cause fatigue and ergonomic strain on the operator, reducing efficiency over time, especially when dealing with cumbersome components. 
\rev{HRC} painting offers a way to mitigate such side effects by employing a robot to assist the operator during the task. 
In this context, we consider a scenario in which the operator is responsible for painting a workpiece, while the robot holds and maneuvers the component to ease the operator's actions. 
Our goal is to devise a motion planning strategy for the robotic assistant that incorporates the operator’s preferences and helps increasing the ergonomics of the painting process. 
Since these preferences are not easily captured through an analytical objective function, we propose to address them by leveraging explicit feedback from the operator after completing the task. 
The proposed approach must meet a set of requirements that consider both \rev{ergonomics} and the overall effectiveness of the operation. 

\begin{requirement}\label{requirement:scaled-up rotations}
    Small changes in orientations of the operator's hand should be associated to \emph{scaled-up} rotations of the piece to be painted. 
    At the same time, the movements should comply with the robot's physical limits.
\end{requirement}

\begin{requirement}\label{requirement:reactiveness}
    To obtain an acceptable surface painting, the robot should be \emph{reactive} in tracking changes in the operator's hand orientation, favoring the regions of the component that the human wishes to paint. 
\end{requirement}

\begin{requirement}\label{requirement:teaching}
    The robot behavior should resemble that of a second operator holding the workpiece and adapting to the first operator's movements. 
    Hence, the trajectory should be generated so that the robot moves in a \emph{human-like} style.
\end{requirement}

\section{Method}\label{sec: method}
Our approach can be summarized as follows. 
We propose a modified version of \rev{DMPs} to generate the robot’s trajectories, where the modification introduces a set of hyperparameters that influence the quality and comfort of the collaborative operation. 
Since manually tuning these parameters to suit each operator's preferences is impractical, we rely on the GLISp algorithm \cite{bemporad2021global} that implements \rev{PBO}. 
This method iteratively adjusts the parameters based on the operator’s feedback, ultimately allowing the system to be fine-tuned to individual preferences. 
This \rev{Section} describes the DMPs-based motion generation (\rev{\ref{sec:DMPs}}), our changes to the standard DMPs (\rev{\ref{sub:coupling terms}}), and the PBO framework (\rev{\ref{sec:PBO}}) \rev{we adopted}.

\subsection{Generation of the robot trajectories}
\label{sec:DMPs}
Dynamic Movement Primitives (DMPs) allow to encode trajectories as solutions of a second-order dynamical system, modulated by a non-linear forcing term learned from a single demonstration. 
Their inherent flexibility allows online modulation of both timing and shape of the trajectory, making them particularly suitable for applications where operational parameters vary during the execution, and the trajectories must be recomputed at run-time.

The standard formulation of DMPs \cite{ijspeert2002movement} does not suit well to the requirements described in Section \ref{sec:problem_statement}, since they do not depend on the process parameters acquired from the environment. 
For this reason, we resort to a variant of the DMPs formulation. 
First, we use a decoupled \rev{set} of equations to generate the robot reference trajectories, for Cartesian positions and orientation. 
This strategy enables the robot to dynamically respond to the operator's actions and adjust the orientation of the piece to be painted accordingly. Because the reference trajectories are taught through teleoperation, the Cartesian position coordinates $\{x,y,z\}$ inherently capture \emph{human-like} behavior (Requirement \ref{requirement:teaching}).
The decoupled set of equations we implemented is as follows:
\begin{equation}
    \begin{aligned}
        \dot{\mathbf{z}}/\bar{\tau} &= \alpha_z \left(\beta_z (\mathbf{g}_p - \mathbf{p}) - \mathbf{z} \right) + \mathbf{f_p}(s) \\
        \dot{\mathbf{p}}/\bar{\tau} &= \mathbf{z} \\
        \boldsymbol{\dot{\eta}}/\bar{\tau} &= \alpha_z \left(\beta_z 2 \log (\mathbf{\bar{g}_o} \otimes \mathbf{q}^*) - \boldsymbol{\eta} \right) + \mathbf{f_o}(s) + \boldsymbol{\gamma}_s \\
        \dot{\mathbf{q}}/\bar{\tau} &= \frac{1}{2} \boldsymbol{\eta} \otimes \mathbf{q} \\
        \dot{s}/\bar{\tau} &= - \alpha_s s
    \end{aligned}
\label{eq:final_system}
\end{equation}
where $\mathbf{z}$ is the \emph{transformation variable} acting as a scaled velocity, while $s$ is the \emph{phase variable}.
We denote by \rev{$\mathbf{p}=[p_x, p_y, p_z]^{\top}$} the Cartesian position of the robot, while the Cartesian goal $\mathbf{g_p}$ is fixed. 
The orientation equations leverage quaternion-based DMPs \cite{saveriano2023dynamic}, \cite{ude2014orientation}: $\mathbf{q} = [q_x, q_y, q_z, q_w]^{\top}$ is the quaternion representing the robot orientation in time; $\mathbf{q^*}$ is the quaternion conjugate, and ``$\otimes$'' is the quaternion product. 
$\boldsymbol{\eta}$ is the \emph{augmented} angular velocity $\boldsymbol{\omega}$, with null scalar part (\rev{$\boldsymbol{\eta} = [\omega_x, \omega_y, \omega_z, 0]^{\top}$}), while $\log(\mathbf{q}):\ \mathbb{S}^3 \rightarrow \mathbb{R}^3$ is the quaternion \emph{logarithmic mapping} that allows to extract the rotation $\mathbf{r} \in \mathbb{R}^3$ (\emph{Rodrigues} representation) from $\mathbf{q}$ \cite{bruno2010robotics}. 
Coefficients $\alpha_z$, $\beta_z$ and $\alpha_s$ are selected to obtain a \emph{critically damped} behavior (see Table \ref{table:user_defined_parameters}), while the weights $\mathbf{w_p}$ and $\mathbf{w_o}$ of the non-linear forcing terms $\mathbf{f_p}(s)$ and $\mathbf{f_o}(s)$ are learned leveraging \emph{locally weighted regression} (LWR) exploiting the general definition presented in \cite{ijspeert2002movement}. 
Finally, we define with $\mathbf{y} = [\mathbf{p}, \mathbf{r}] \in \mathbb{R}^6$ the complete state of the system and with $\mathbf{\dot{y}, \mathbf{\ddot{y}}}$ the velocity and the acceleration, respectively. 
To account for the modifications described in Section \ref{sec:problem_statement} we introduced the corrective terms $\boldsymbol{\gamma}_s$, $\mathbf{\bar{g}_o}$ and $\bar{\tau}$, \rev{detailed} in the following.

\subsection{Explanation of the DMPs corrective terms}
\label{sub:coupling terms}
Requirement \ref{requirement:scaled-up rotations} is related to the \emph{feasibility} of the task. 
Especially for cumbersome pieces, confining the range of motion of the operator's hand in a limited space leads to \rev{reduced efforts} during the painting process \cite{zanchettin2019collaborative}. 
To address this requirement in a systematic way, we introduced the reference frames in Fig. \ref{fig:setup_reference_frames}, all expressed with respect to the robot base frame $\mathbf{\{h_B\}}$. 
We denoted by $\textbf{\{$\mathbf{h_0}$\}}$ the reference frame associated to the operator's hand in the first time instant of the process, which is fixed with respect to the robot base. 
During the execution of the task, the orientation of the human hand, expressed with $\textbf{\{$\mathbf{h_H}$\}}$, is tracked by the camera whose frame is denoted as $\textbf{\{$\mathbf{h_C}$\}}$. 
Based on this, the relative rotation of the hand with respect to $\textbf{\{$\mathbf{h_0}$\}}$ can be computed in terms of rotation matrices as follows:

\begin{figure}
\centering
\includegraphics[width=0.45\textwidth]{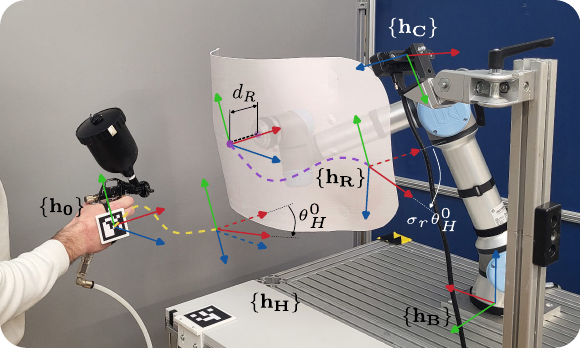}
\caption{Setup used for painting. The yellow and purple dashed lines represent the human and robot trajectories, respectively.}
\label{fig:setup_reference_frames}
\end{figure}

\begin{equation}
\mathbf{R_H^0} = (\mathbf{R_0^B})^{-1}\mathbf{R_H^B}
\label{eq:relative orientation}
\end{equation}
Equation \eqref{eq:relative orientation} can easily be transformed to $\mathbf{r_H^0}= \theta_H^0 \mathbf{u_H^0}$ leveraging the \emph{Rodrigues} notation ($\theta_H^0$ and $\mathbf{u_H^0}$ can be computed as described in \cite{bruno2010robotics}). 
To address Requirement \ref{requirement:scaled-up rotations} we compute the \emph{scaled} rotation as follows:
\begin{equation}
  \mathbf{\bar{r}_H^0} = \sigma_r  \theta_H^0 \mathbf{u_H^0},\quad \text{ with }  \sigma_r = 1 + \frac{k_m - 1}{1 + e^{-a_d (||\mathbf{r_H^0}|| - \delta_d)}}
\label{eq: sigma_r}
\end{equation}
where $\sigma_r$ represents a decreasing sigmoidal function that can take values between $k_m$, that is a tunable parameter, and 1. 
Parameters $a_d$ and $\delta_d$ are the steepness and the center of the sigmoidal, respectively. 
In this way, after computing the equivalent matrix form of equation \eqref{eq: sigma_r} ($\mathbf{\bar{R}_H^0}$), it is possible to invert equation \eqref{eq:relative orientation} to find $\mathbf{\bar{R}_H^B}$. 
Finally, \rev{to keep} consistency with equation \eqref{eq:final_system}, $\mathbf{\bar{R}_H^B}$ is transformed into $\mathbf{\bar{g}_o}$, which represents the scaled absolute rotation of the operator's hand that should be reached by the robot.
Referring to Fig. \ref{fig:setup_reference_frames}, the robot is supposed to rotate \rev{by} the \emph{scaled} quantity $\sigma_r \theta_H^0$ as a response to a rotation of $\theta_H^0$ of $\mathbf{\{h_H\}}$.
For how we engineered our framework, the $\{x,y,z\}$ Cartesian coordinates of the human hand do not influence the performance, since our main goal is to reduce the operator's hand range of motion. 
Despite DMPs would also allow to implement additional modifications, we decided to adjust the robot's behavior only \rev{in terms of} orientation, allowing the operator to freely modulate their Cartesian distance to the \rev{workpiece}.
\rev{As shown in Section \ref{sec:experiments}, this choice does not preclude the possibility of applying the framework with operators characterized by heterogeneous heights, even though the $z$ Cartesian coordinates are not directly considered}.

Requirement \ref{requirement:reactiveness} concerns the \emph{responsiveness} of the robot with respect to changes in the orientation of the goal (\rev{\emph{i.e.,}} the human's hand). To account for this, we introduced
\begin{equation}
\boldsymbol{\gamma}_s = \alpha_z k_s \sigma_s \boldsymbol{\eta}, \quad\text{ with } \sigma_s = \frac{1}{1 + e^{a_s(\text{d}(\mathbf{\bar{g}_o}, \mathbf{q}) - \delta_s)}}
\label{eq:gamma_s}
\end{equation}
where $k_s$ is a tunable scaling factor, while $\sigma_s$ is a sigmoidal function centered in $\delta_s$ and with steepness $a_s$. 
The term $\text{d}(\mathbf{\bar{g}_o}, \mathbf{q}) \in \text{S}^3$ is defined as in \cite{ude2014orientation} and represents a \emph{distance} metric valid for rotations. 
The rationale behind $\boldsymbol{\gamma_s}$ is the following: in case the relative rotation between human and robot is above a limit ($\delta_s$), the robot should quickly \emph{react} to this mismatch, trying to catch up with the orientation of the operator (possibly scaled according to equation \eqref{eq: sigma_r}). 
To fully exploit this behavior, the product $k_s \sigma_s$ should be as close as possible to 1: in this case, as visible from equation \eqref{eq:final_system}, the term $\boldsymbol{\gamma}_s$ would be equal and opposite to $\alpha_z\boldsymbol{\eta}$, removing the damping from the system and increasing the responsiveness. 
\rev{For this reason though}, an upper bound on $k_s$ needs to be selected to avoid that the system becomes unstable. 

The standard DMPs formulation \cite{ijspeert2002movement} explicitly uses the trajectory execution time, $\tau$. 
On the contrary, we use $\bar{\tau} = 1/\tau$, where $\bar{\tau}$ can be regarded as a time scaling factor, ranging from 0 (infinite execution time) to 1 (execution time \rev{equal to 1} second).
The reason is that the PBO algorithm we exploited (see Section \ref{sec:PBO}) is a learning-based iterative scheme and, to be effective, the design variables should have similar variational ranges, to balance their influence during the process.

Based on $\bar{\tau}$ and the scaling factors $k_m$ and $k_s$ introduced with equations \eqref{eq: sigma_r} and \eqref{eq:gamma_s}, a set of tunable parameters emerges that directly influence the robot’s responsiveness and its adaptive behavior. 
To optimize these parameters according to individual user preferences, we integrate the PBO framework presented in Section \ref{sec:PBO}.

\subsection{Optimization of the process parameters}
\label{sec:PBO}
In the context of HRC painting, the underlying \textit{objective function} representing the process is difficult to model explicitly, considering the complexity and high dimensionality of the design variables involved. 
By defining $\boldsymbol{\theta} \in \mathbb{R}^{n_{\theta}}$ as the set of parameters and $J(\boldsymbol{\theta}) : \mathbb{R}^{n_{\theta}} \rightarrow \mathbb{R}$ as the objective function, we can formalize the general initial problem as follows:
\begin{equation}
\begin{aligned}
\boldsymbol{\theta^*} = \arg \min_{\boldsymbol{\theta}} \quad & J(\boldsymbol{\theta})\\
\textrm{s.t.} \quad & \boldsymbol{\theta_l} \leq \boldsymbol{\theta} \leq \boldsymbol{\theta_u}\\
  &\boldsymbol{\theta}\in \Gamma    \\
\end{aligned}
\label{eq:starting min problem}
\end{equation}
where $\boldsymbol{\theta_l}$ and $\boldsymbol{\theta_u}$ are the lower and upper limits on the domain, while $\Gamma : \{\boldsymbol{\theta} \in \mathbb{R}^{n_{\theta}} : g(\boldsymbol{\theta}) \leq 0\}$ is the set of $q$ inequalities $g : \mathbb{R}^{n_{\theta}} \rightarrow \mathbb{R}^q$ constraining the problem.

To overcome the complexity of modeling $J$, we solve \eqref{eq:starting min problem} through PBO by using the GLISp algorithm \cite{bemporad2021global}. 
The PBO paradigm works as follows: it samples a candidate vector, and a new test implementing such parameters is executed. 
At the end of the test, a subject is asked to evaluate whether the last execution was \rev{\emph{better}}, \rev{\emph{as good as}}, or \rev{\emph{worse}} than the previous execution. 
The procedure is repeated for $N$ tests, provided that this number guarantees convergence.
Consequently, the problem of minimizing an objective function $J$ can be interpreted as finding $\boldsymbol{\theta^*}$ such that $\pi(\boldsymbol{\theta^*}, \boldsymbol{\theta}) \leq 0, \forall \boldsymbol{\theta} \in \Gamma,\  \boldsymbol{\theta_l} \leq \boldsymbol{\theta} \leq \boldsymbol{\theta_u}$ after defining the following \textit{preference function} $\pi :  \mathbb{R}^{n_{\theta}} \times  \mathbb{R}^{n_{\theta}} \rightarrow \{-1, 0, 1\}$: 
\begin{equation}
\pi(\boldsymbol{\theta_i}, \boldsymbol{\theta_j}) =
\begin{cases} 
    -1 & \text{if } \boldsymbol{\theta_i} \text{ is ``better'' than } \boldsymbol{\theta_j} \\
     0 & \text{if } \boldsymbol{\theta_i} \text{ is ``as good'' \rev{as} }  \boldsymbol{\theta_j} \\
     1 & \text{if } \boldsymbol{\theta_i} \text{ is \rev{``worse''} than } \boldsymbol{\theta_j}
\end{cases}
\label{eq:preference function}
\end{equation}
Consider a set of $N \geq2$ samples, $\Theta = \{\boldsymbol{\theta_1}, \boldsymbol{\theta_2}, \cdots, \boldsymbol{\theta_N}\}$, with $\boldsymbol{\theta}_i, \boldsymbol{\theta}_j \in \mathbb{R}^{n_{\theta}}$ and such that $\boldsymbol{\theta}_i \neq \boldsymbol{\theta}_j \forall i\neq j$, with $i,j = 1, \cdots, N$.
\rev{Based on $\Theta$,} a \textit{preference vector} $\rev{\mathbf{B} = [b_1, \cdots, b_M]^{\top}} \in \{-1, 0, 1\}^M$ can be created \rev{through} equation \eqref{eq:preference function}.
Here, $M$ is the number of preferences and such that $1 \leq M \leq \binom{N}{2}$. Each element $b_h = \pi(\boldsymbol{\theta}_{i(h)}, \boldsymbol{\theta}_{j(h)})$, with $h \in \{1, \cdots, M\}$, $i(h), j(h) \in \{1, \cdots, N\}$ and $i(h) \neq j(h)$, encodes the preference with respect to the experiment $h$.

Since the underlying function $J$ is unknown, we substitute it with a surrogate function $\hat{J}:\mathbb{R}^{n_{\theta}} \rightarrow \mathbb{R}$ featured by a preference vector $\hat{\pi}$, defined in the same way as equation \eqref{eq:preference function} and such that $\pi(\boldsymbol{\theta}_{i(h)}, \boldsymbol{\theta}_{j(h)}) = \hat{\pi}(\boldsymbol{\theta}_{i(h)}, \boldsymbol{\theta}_{j(h)})$, $\forall h \in \{1, \cdots, M\}$. Based on the work proposed in \cite{gutmann2001radial}, we defined the surrogate $\hat{J}$ as the following \rev{\emph{radial basis function}} (RBF):
\begin{equation}
    \hat{J}(\boldsymbol{\theta}) = \sum_{i=1}^N \beta_i \phi(\epsilon d_e(\boldsymbol{\theta}, \boldsymbol{\theta}_i))
\label{eq:surrogate_function}
\end{equation}
where $\epsilon > 0$ is a scalar hyper-parameter, $d_e(\boldsymbol{\theta}_i, \boldsymbol{\theta}_j) = \| \boldsymbol{\theta}_i - \boldsymbol{\theta}_j\|^2, \boldsymbol{\theta}_i, \boldsymbol{\theta}_j \in \mathbb{R}^{n_{\theta}}$ represents the Euclidean distance, while $\phi:\mathbb{R}\rightarrow\mathbb{R}$ is the actual RBF, that we defined as $\phi(\epsilon d) = \frac{1}{1 + (\epsilon d)^2}$ (\emph{inverse quadratic} definition). 
The vector \rev{$\boldsymbol{\beta} = [\beta_1, \cdots, \beta_N]^{\top}$} contains the coefficients $\beta_i$ appearing in equation \eqref{eq:surrogate_function}, which can be computed by solving the convex optimization problem proposed in \cite{bemporad2021global}.

\rev{To enhance performances}, we introduced a \emph{recalibration} strategy for the hyper-parameter $\epsilon$, similar to the one implemented in \cite{campagna2024promoting}. For this purpose, we define a sequence of $L$ elements $\boldsymbol{\alpha} = \{10^{-1+\frac{1}{5}(l-1)}\}_{l=1}^{L=10}$ and, when the recalibration is triggered, $L$ new values of $\epsilon$ are computed (denoted as $\bar{\boldsymbol{\epsilon}}$): the selected one ($\epsilon^*$), minimizes the \emph{leave-one-out cross-validation error}. By defining with $\bar{N}$ the number of initial samples for the training phase, the set containing the indices of the experiments requiring recalibration can be computed as 
\begin{equation}
\mathcal{I} = \{\bar{N}, \bar{N} + \lfloor \tfrac{N - \bar{N}}{4} \rfloor, \bar{N} + \lfloor \tfrac{N - \bar{N}}{2} \rfloor, \bar{N} + \lfloor \tfrac{3(N - \bar{N})}{4} \rfloor\}
\label{eq:I_set}
\end{equation}

For an effective sampling of the search space, a surrogate function $\hat{J}$ is usually not enough. To overcome this issue, an acquisition function $a(\boldsymbol{\theta})$ can be used, provided that the problem proposed in equation \eqref{eq:starting min problem} is reformulated as an iterative scheme by which new candidates are proposed:
\begin{equation}
\boldsymbol{\theta}_{k+1} = \arg \min_{\underset{\boldsymbol{\theta}_l \leq \boldsymbol{\theta} \leq \boldsymbol{\theta}_u}{\boldsymbol{\theta} \in \Gamma}} a(\boldsymbol{\theta})
\label{eq:final PBO}
\end{equation}
where $k$ represents the current iteration. In our work, we defined the acquisition function as:
\begin{equation}
    a(\boldsymbol{\theta}) = \frac{\hat{J}(\boldsymbol{\theta})}{\max_i(\hat{J}(\boldsymbol{\theta_i}))  - \min_i(\hat{J}(\boldsymbol{\theta_i}))} - \delta_E z_N(\boldsymbol{\theta})
\label{eq:acquisition_function}
\end{equation}
where $\boldsymbol{\theta}$ is the space limited by $\boldsymbol{\theta}_l$ and $\boldsymbol{\theta}_u$, $\boldsymbol{\theta_i}$ is a generic sample in the domain, $\delta_E\geq 0$ is the \emph{exploration parameter}, while $z_N(\boldsymbol{\theta})$ is the 
exploration function, defined as in \cite{zhu2021c}.

\rev{Alg. \ref{alg:PBO} shows the complete procedure}. In steps \ref{step: inversion 1}, \ref{state: training} and \ref{step: inversion 2}, we denote by $\pi^{-1}(\mathbf{B}(k))$ the operation that allows to retrieve the specific element $\Theta(k)$ that resulted in $\mathbf{B}(k)$. 

\section{Experimental validation and use case}
\label{sec:experiments}
We consider a Universal Robots UR5e cobot, equipped with a 3D-printed support to keep the piece at $d_{\text{R}}=50$ mm from the flange (see Fig. \ref{fig:setup_reference_frames}). 
All algorithms ran on a laptop with 16 GB of RAM and with a Intel(R) Core(TM) i7-10750H CPU @ 2.60GHz. 
We used ROS Noetic to handle the communication protocols and to collect the data (the sets $\mathcal{X}_i$ and $\mathcal{Z}_i$ described in Subsection \ref{sub:ergonomy for comfort}) coming from a OAK-D Pro camera at 30 Hz. 
Table \ref{table:user_defined_parameters} shows the selected parameters.

\begin{figure}
\centering
\includegraphics[width=0.425\textwidth]{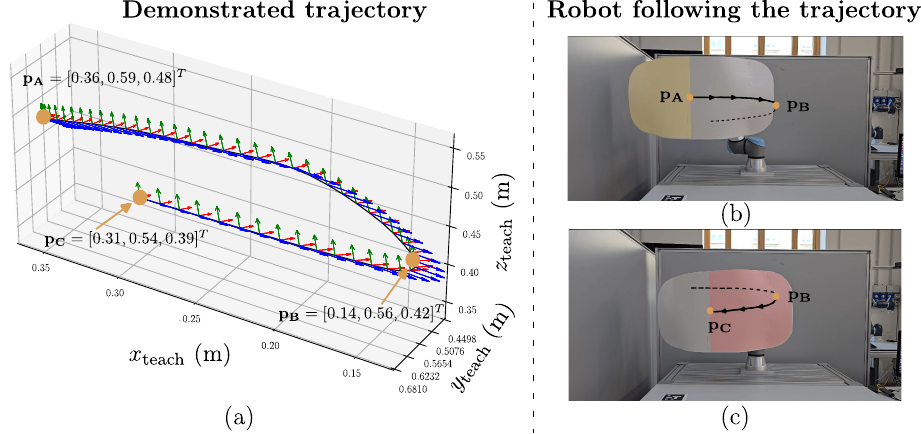}
\caption{(a) Demonstrated trajectory, taught offline with teleoperation and associated to $\mathbf{\{h_R\}}$. (b) Robot in $\mathbf{p_A}$. (c) Robot in $\mathbf{p_C}$.}
\label{fig:reference_frames_trajectory}
\end{figure}

The experiments \rev{consist} of $N_p=15$ tests, each performed by a different participant, denoted as $\text{P}i$, with $i \in \{1,\cdots,15\}$. 
The set was composed of 8 men and 7 women among PhD and master's students, with an average age of 25.2 years with standard deviation of $\pm 0.98$ years, and an average height of 180.8 cm with standard deviation of $\pm 11.26$ cm.
As shown in Fig. \ref{fig:reference_frames_trajectory}a, the baseline trajectory taught with teleoperation (applying minor modifications to the work presented in \cite{welle2024quest2ros}) presents a \emph{c-shaped} profile, and is kept the same for all participants to obtain comparable results. 
\rev{We preferred this strategy over more intuitive ones, such as tracking the pose of a marker on the workpiece, because it provides more stable and reliable pose estimates, avoiding the noise and flickering that can affect vision-based measurements.}
The painting strategy can be seen \rev{on the right side of Fig. \ref{fig:reference_frames_trajectory}}: as the robot moves from point $\mathbf{p_A}$ to point $\mathbf{p_B}$ (\rev{Fig.} \ref{fig:reference_frames_trajectory}b), the operators should paint the part of the piece on their left (yellow); the other part of the piece (red) should be painted as the robot moves from point $\mathbf{p_B}$ to point $\mathbf{p_C}$ (Fig. \ref{fig:reference_frames_trajectory}c). 
Moreover, the component we used in the process is characterized by an extended surface (\rev{$\approx 1\;m^2$}) and is featured with different curvature radii, which entail challenging painting conditions for the operators and a meaningful testbed for our framework.
We substituted the real paint with natural coloring for food, which is non-toxic and does not require specific ventilation devices, but allowed us to perform \emph{visual inspection} of the results considering the percentage painted surface as the main reference parameter \cite{atkar2005uniform}. 
A video of the experiments is available at this link\footnote{Video of the experiments: \url{https://youtu.be/0POZMUFvQUE}}.

The experiments consisted of two tests. 
In the first one, each participant was asked to paint the piece while the robot was kept still in $\mathbf{p_A}$. 
This first test is representative of a non-collaborative baseline, common in 
current industrial setups.
In the second test, we implemented the procedure summarized with \rev{Alg.} \ref{alg:PBO}. 
Each complete experiment ($N$ iterations and the static case) lasted approximately 1 hour, due to the setup and cleaning times between experiments. 

We aim at optimizing \rev{$\boldsymbol{\theta} = [\bar{\tau}, k_s, k_m]^{\top}$} ($n_{\theta}=3$), 
since they are the main parameters characterizing the robot behavior modeled through equation \eqref{eq:final_system}.
We set the upper and lower limits as \rev{$\boldsymbol{\theta_u}= [0.1, 0.7, 1.5]^{\top}$} and \rev{$\boldsymbol{\theta_l}=[0.02, 0.3, 1.0]^{\top}$}.

\begin{algorithm}[tpb]
\scriptsize
\caption{Complete painting process (PBO + DMPs)}
\label{alg:PBO}
\begin{algorithmic}[1]

    \STATE \textbf{Given}: $N$, \rev{$\bar{N}$}, $n_{\theta}$, $\mathcal{I}$, $\boldsymbol{\alpha}$, $\boldsymbol{\theta}_l \leq \boldsymbol{\theta} \leq \boldsymbol{\theta}_u$
    \STATE \textbf{Initialize}: $\mathbf{B}\gets\mathbf{0}$, $\boldsymbol{\beta}\gets \mathbf{0}$, $\Theta \gets \mathbf{0}$, $\boldsymbol{\theta}^* \gets \mathbf{0}$, $\epsilon\gets 1$
    \STATE $\Theta(1:\bar{N}) \gets \rev{\texttt{GenerateInitialSamples}}(\boldsymbol{\theta_l}, \boldsymbol{\theta_u})$  \label{step: latin_sampling}
    \STATE $\texttt{ModifiedDMPs}$($\Theta(1), \Theta(2)$), \rev{$\texttt{ExecPainting}$($\Theta(1), \Theta(2)$})
    \STATE $\mathbf{B}(1) \gets \pi(\Theta(1), \Theta(2))$, $\boldsymbol{\theta}^* \gets \pi^{-1}(\mathbf{B}(1))$ \label{step: inversion 1}
    \FOR{$i \gets 2,\dots, N$}
    \STATE $\texttt{ModifiedDMPs}$($\Theta(i+1)$), \rev{$\texttt{ExecPainting}$($\Theta(i+1)$)} 
        \IF {$i < \bar{N}$}
            \STATE $\mathbf{B}(i) \gets \pi(\Theta(i+1), \boldsymbol{\theta}^*)$, $\boldsymbol{\theta}^* \gets \pi^{-1}(\mathbf{B}(i))$ \label{state: training}
        \ELSE
            \IF{$i \in \mathcal{I}$ \rev{($\mathcal{I}$ computed with Eq. \eqref{eq:I_set})}}
                \STATE $\boldsymbol{\bar{\epsilon}} \gets \boldsymbol{\alpha} \epsilon$, $\epsilon^* \gets \texttt{Recalibration}$($\boldsymbol{\bar{\epsilon}}$), $\epsilon \gets \epsilon^*$
            \ENDIF
            \STATE $\mathbf{B}(i) \gets \pi(\Theta(i), \boldsymbol{\theta}^*)$, $\boldsymbol{\theta}^* \gets \pi^{-1}(\mathbf{B}(i))$ \label{step: inversion 2}
            \STATE $\boldsymbol{\beta} \gets \texttt{SolveConvexProblem}(\mathbf{B}(i))$
            \STATE $\hat{J}(\boldsymbol{\theta}) \gets \sum_{j=1}^i{\boldsymbol{\beta}(j) \phi(\epsilon d(\boldsymbol{\theta},\Theta(j)))}$ \rev{(Eq. \eqref{eq:surrogate_function})}
            \STATE $a(\boldsymbol{\theta}) \gets \frac{\hat{J}(\boldsymbol{\theta})}{\max_k(\hat{J}(\boldsymbol{\theta_k}))  - \min_k(\hat{J}(\boldsymbol{\theta_k}))} - \delta_E z_N(\boldsymbol{\theta})$ \rev{(Eq. \eqref{eq:acquisition_function})}
            \STATE $\Theta(i+1) \gets \texttt{ParticleSwarmOptimization}(a(\boldsymbol{\theta}))$ \label{step: PSO} \rev{(Eq. \eqref{eq:final PBO})}
        \ENDIF
    \ENDFOR
    \STATE \textbf{return}: $\boldsymbol{\theta}^*$
\end{algorithmic}
\end{algorithm}

\begin{table}[h]
\caption{\rev{User-defined} parameters for DMPs and PBO.}
    \centering
    \small 
    \resizebox{0.46\textwidth}{!}{ 
    \begin{tabular}{ccccccc|ccc}
        \toprule
        \multicolumn{7}{c|}{\textbf{DMPs}} & \multicolumn{3}{c}{\textbf{PBO}}\\
        \midrule
        $a_d$ & $\delta_d$ & $a_s$ & $\delta_s$ & $\alpha_s$ & $\alpha_z$ & $\beta_z$ & $\bar{N}$ & $N$ & $\delta_E$ \\
        \toprule
        -10 & 0.35 rad & -10 & 0.35 rad & 5 & 25 & 6.25 & 6 & 15 & 0.5\\
        \bottomrule
    \end{tabular}
    } 
    \label{table:user_defined_parameters}
\end{table}

\subsection{Quantitative metrics for ergonomics assessment}
\label{sub:ergonomy for comfort}
According to the painting strategy, we expect a reduction in the range of motion of the operator's hand. 
In the literature, several scores have been proposed with the aim of quantifying the overall ergonomics \cite{kee2022systematic}, and they are meant to be applicable for a broad class of tasks. 
For the specific settings we tackle, we preferred to propose two custom indicators that were \rev{tailored to} collaborative painting. Indeed, in this task the human is mainly required to move their hand along the $x$ and $z$ axis of $\mathbf{\{h_B\}}$, while the other body parts are kept roughly still. For this reason, traditional ergonomics scores would most likely provide estimations of the overall comfort characterized by lower reliability than the proposed ones, since they may struggle in catching process-specific criticalities.

\begin{figure}
\centering
\includegraphics[width=0.45\textwidth]{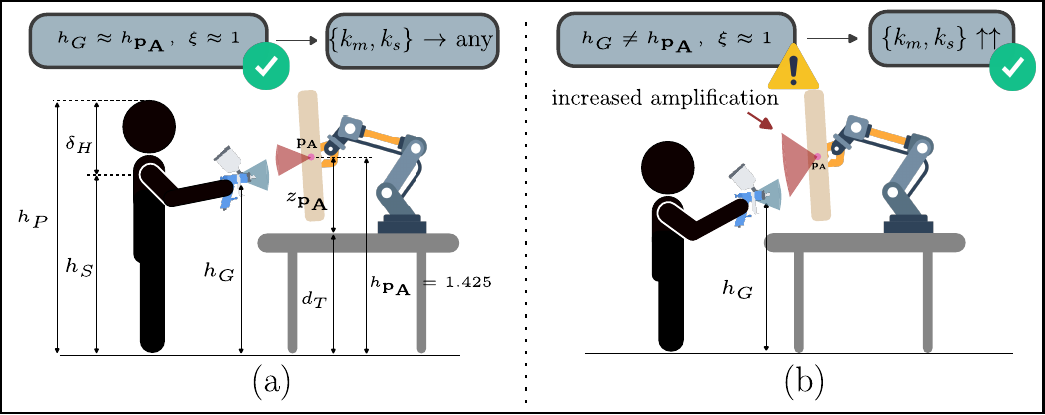}
\caption{(a) $h_G \approx h_S \approx h_{\mathbf{p_A}}$: comfortable starting position. (b) Acceptable \rev{case}: the operator will rely on bigger values of $k_m$ and $k_s$ if $h_S \neq h_{\mathbf{p_A}}$.}
\label{fig:shoulder_problem}
\end{figure}

For what concerns the height (\emph{overhead} ergonomics), we considered the distances shown in Fig. \ref{fig:shoulder_problem}a. 
For the $i$th participant we computed the ratio $\xi_i = h_{G,i}/h_{S,i}$, where $h_{S,i}$ is the shoulder height with respect to the floor, that we evaluated as $h_{S,i} = h_{P,i} - \delta_H$ ($h_{P,i}$ is the height of the participant, while we used an average head height $\delta_H = 0.3$ m). 
Instead, $h_{G,i}$ represents each element of the set $\mathcal{Z}_i$ containing the $z$ coordinates of the human hand. 
We evaluate the hand-to-shoulder height ratios obtained with the optimized parameters, relative to that obtained in the static case. To do so, we define

\begin{equation}
    \lambda_{z, i} = \frac{\sigma(\xi_{i, \text{static}}) - \sigma(\xi_{i, \text{opt}})}{\sigma(\xi_{i, \text{static}})}
\label{eq:lambdaz}
\end{equation}
where $\sigma$ is the standard deviation of the collected data and $\xi_{i, \text{static}}, \xi_{i, \text{opt}}$ are the ratios $\xi$ for the $i$th participant in the static test and the optimized tests, respectively.
We considered the optimization procedure to be successful in improving \emph{overhead} ergonomics if $\xi_i < 1$ (hand below the shoulder level) and $\lambda_z > 0$ (reduction of the hand variability in $z$).

We introduced the \rev{ratio} $\xi$ in our analysis since it also presents a direct connection to the optimization variables contained in $\boldsymbol{\theta}$. 
Before starting, each participant was asked to position their hand as close as possible to $\mathbf{p_A}$ ($\mathbf{\{h_0\}} \approx \mathbf{\{h_R\}}(t_0)$). 
As shown in Fig. \ref{fig:shoulder_problem}a, for the $z$ coordinate, this is equivalent to asking that $h_{G,i} \approx h_{\mathbf{p_A}}$, always satisfying the requirement $\xi_i < 1$. 
On the contrary, $h_{G,i} \approx h_{\mathbf{p_A}}$ and $\xi \neq 1$ was considered too risky; consequently, an initial condition similar to the one presented in Fig. \ref{fig:shoulder_problem}b was preferred, accepting that $h_{G,i} \neq h_{\mathbf{p_A}}$. 
In this case, we expected the participant to select higher values of $k_m$ and $k_s$ to compensate for the mismatch between the hand pose and the \rev{work-piece center}. 

For the $x$ coordinate (\emph{lateral} ergonomics), we considered the standard deviation of the values of $x_H$. 
Based on the two regions that are created in the optimal case (see the video at minute 02.11), the set of the $x$ coordinates of the hand positions $\mathcal{X}_i$ is divided into $k=2$ clusters ($\mathcal{X}_{i,1}$ and $\mathcal{X}_{i,2}$) exploiting \emph{k-means clustering}, and it is compared to the same set obtained in the static case ($\mathcal{X}_{i, \text{static}}$) by introducing:

\begin{equation}
    \lambda_{x, i} = \frac{\sigma(\mathcal{X}_{i, \text{static}}) - \max(\sigma(\mathcal{X}_{(i,1), \text{opt}}), \sigma(\mathcal{X}_{(i,2), \text{opt}}))}{\sigma(\mathcal{X}_{i, \text{static}})}
\label{eq:lambdax}
\end{equation}

\rev{In the end}, we assumed the optimization procedure to be successful in improving \textit{lateral} ergonomics if $\lambda_{x,i} > 0$.

\subsection{Analysis of the quantitative metrics}
\label{sub:analysis_quantitative}

\begin{figure*}
\centering
\includegraphics[width=0.98\textwidth]{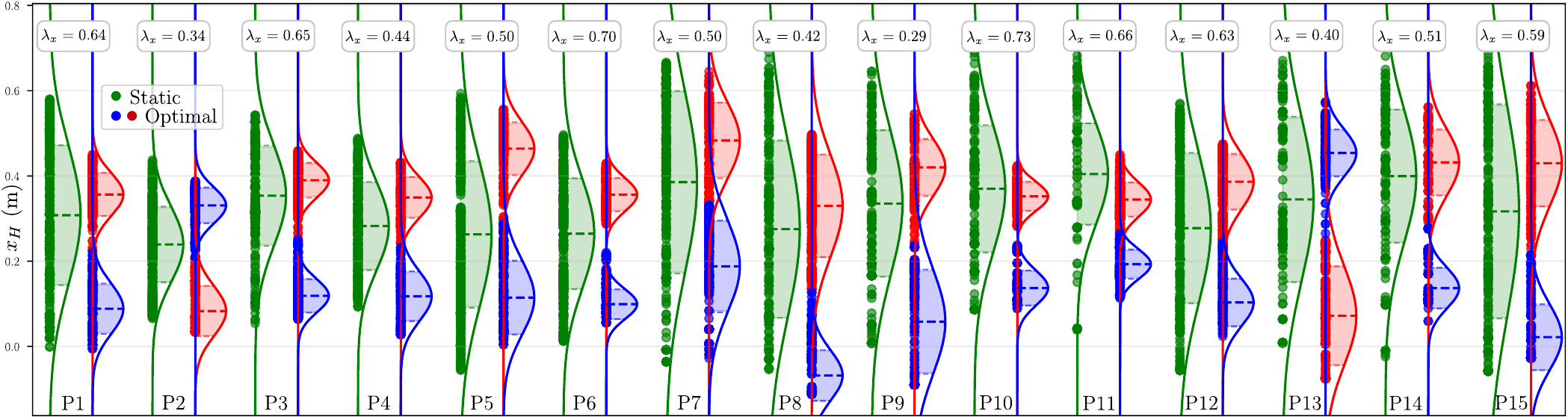}
\caption{Results for the \emph{lateral ergonomics}. For each participant P$i$, the red and blue curves (depicting the optimal case), are obtained through the \emph{k-means} clustering technique. Despite representing the human hand positions along the $x$ axis (horizontal), the data are reported vertically for readability.}
\label{fig:lateral_ergonomics}
\end{figure*}

The results for the \emph{lateral} ergonomics are reported in Fig. \ref{fig:lateral_ergonomics}, which represents the dispersion of $x_H$ data, both in the static case (green curves) and in the optimal ones (represented by the blue and red curves). 
As it can be noticed, the painting effort is successfully reduced for all the participants, since they were able to keep their hand inside two smaller regions with respect to the initial one, that was approximately as extended as the painted component. The maximum reduction for the span of the $x$ coordinates, computed with \eqref{eq:lambdax}, happens for P10 ($\approx 73\%$), while the smallest ratio is obtained by P9, whose hand spread around the mean value is reduced by $\approx 29\%$.

\begin{figure*}
\centering
\includegraphics[width=0.98\textwidth]{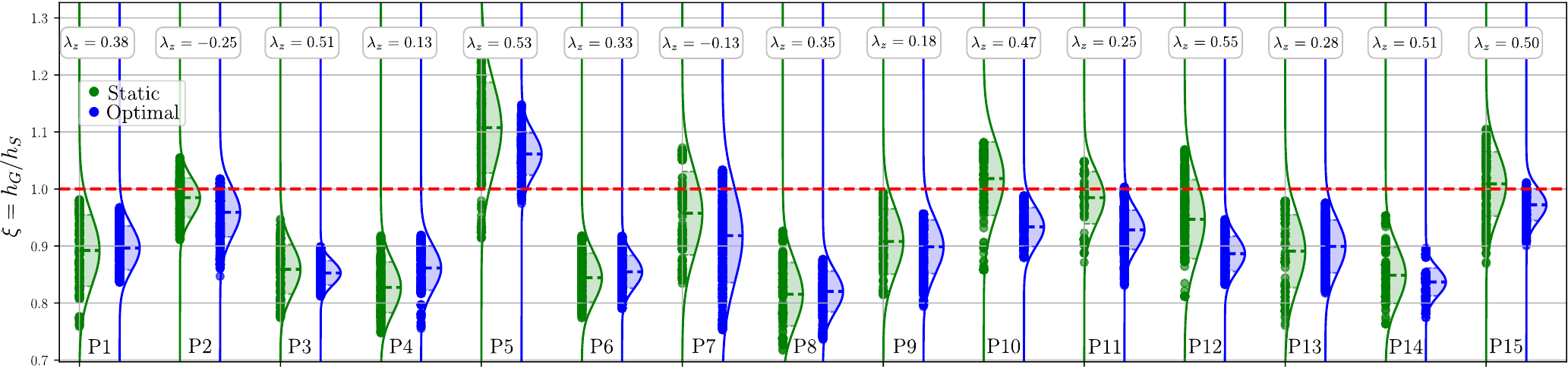}
\caption{Results for the \emph{overhead} ergonomics. The dashed red line represents the acceptable limit on $\xi$, the green area is the data dispersion for the static case, while the blue one represents the optimized situation.}
\label{fig:Gaussians_z}
\end{figure*}

For the \emph{overhead} ergonomics, the results are reported in Fig. \ref{fig:Gaussians_z}, which shows the dispersions associated to the $z$ positions of the human hand in the static case (green) and in the optimal one (blue). On average, our procedure proved effective in obtaining $\xi_i < 1$ and $\lambda_z > 0$ (computed with \eqref{eq:lambdaz}), with a mean reduction in the motion range of $\approx 31\%$. Participant P12 presents the highest reduction in $\lambda_z$ ($\approx 55\%$), while P4 shows the smallest ($\approx 13\%$) ratio. P2 and P7 are the only ones showing an increase in the motion range around the $z$ coordinate ($\approx-25\%$ and $\approx-13\%$, respectively). However, unlike the static case, in the optimal one the hand is never placed above the shoulder level, resulting in an increase in the \emph{overhead} comfort ($\xi_{2,\text{opt}}$ and $\xi_{7,\text{opt}}$ are both smaller than 1), despite the higher variability in the $z$ coordinate. P5 is the only participant for which the procedure, despite reducing the hand variability, is not able to fully guarantee the comfort. 

\begin{figure*}
\vspace{-0.2 cm}
\centering
\includegraphics[width=0.98\textwidth]{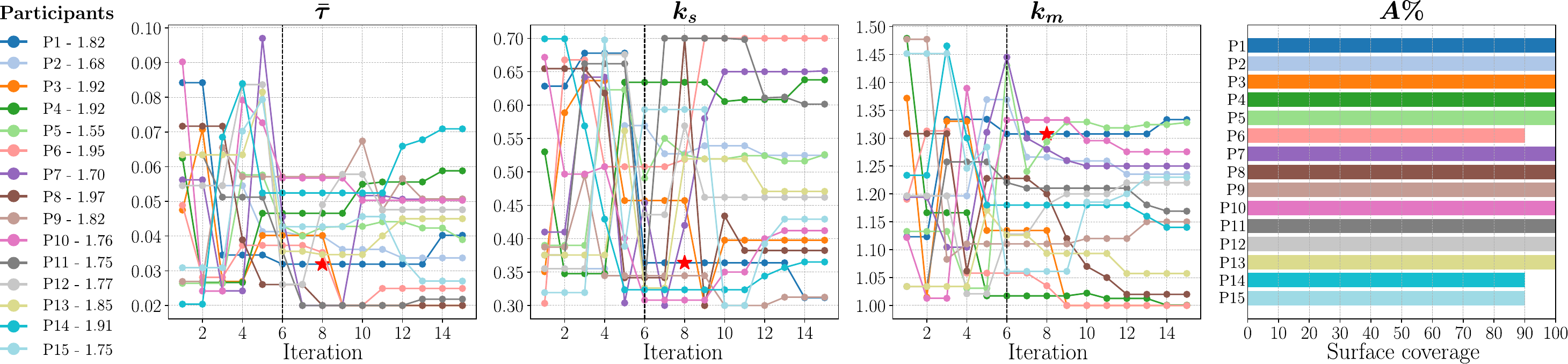}
\caption{Trend of the parameters composing $\boldsymbol{\theta}$ as the number of iterations increases and the percentage painted surface $A\%$ for the final (optimal) test. The dashed vertical line separates the \emph{training} iterations from the \emph{active learning} stage. The point characterized by a red star in the trends for participant 1 at iteration 8 denotes an infeasible operation. In the legend, we also report the values of $h_{P,i}$ in meters.}
\label{fig:PBO_iterations}
\end{figure*}

\subsection{Analysis of the PBO parameters}
\label{sub:analysis_PBO}
The trends of the PBO parameters are visible in Fig. \ref{fig:PBO_iterations}. 
As it can be seen from the plots, the chosen boundaries were broad enough to obtain feasible experiments, in accordance with Requirement \ref{requirement:scaled-up rotations}: only at the 8th \rev{iteration} for P1, indicated with a red star in Fig. \ref{fig:PBO_iterations}, the operation was infeasible, due to too high joint velocities.
For P3, P4, P6 and P8, the values of $k_{m, \text{opt}}$ are quite similar and close to the lower limits: this can easily be explained by the fact that they never have to compensate for an initial mismatch with respect to the center of the piece ($h_S > h_{\mathbf{p_A}}$). 
Instead, P5 settled for values of $k_m$ and $k_s$ that are among the highest, in accordance with the situation presented in Fig. \ref{fig:shoulder_problem}b. 
In general, this seems a common trend, since the preferences of participants characterized by similar heights tend to converge to similar regions, showing the importance of devising a framework where the robot behavior can be tuned according to subjective needs.
In general, the time scaling converges to more subjective values, ranging from \rev{$\approx 14$ to $\approx 50$ seconds}, and each participant settled for a unique value based on their own pace during the operation.

\subsection{Qualitative metrics and perception of the process}
\label{sub:qualitative metrics}
At the end of each experiment, all candidates were asked to express their preference between the set of parameters proposed by PBO for that \rev{iteration} and the best set selected \rev{until that moment}, 
and they were asked to fill out a questionnaire inspired by the questions presented in \cite{lasota2015analyzing}, adapting and expanding them to suit the specific case under analysis. 
The proposed questions were as follows:

\begin{question} \label{question:time}
    \emph{Did the operation last a reasonable amount of time?} ($\mathbf{Q_1}$) 
\end{question}

\begin{question} \label{question:robot assistence}
    \emph{How would you rate the robot assistance?} ($\mathbf{Q_2}$) 
\end{question}

\begin{question} \label{question:ergonomics}
    \emph{How would you rate the general comfort of the operation?} ($\mathbf{Q_3}$) 
\end{question}

The participants were asked to answer with scores ranging from 1 (\emph{not satisfied at all}) to 5 (\emph{completely satisfied}).

\rev{It is important} to point out that the questionnaire and the related statistical analysis are not meant to be an exhaustive evaluation, but aim at providing an additional assessment of the workers’ perceived impact, in order to better contextualize the \rev{obtained results}.

\subsection{Analysis of the questionnaire scores}
\label{sub:analysis_questionnaire}
\begin{figure}
\vspace{-0.1 cm}
\centering
\includegraphics[width=0.47\textwidth]{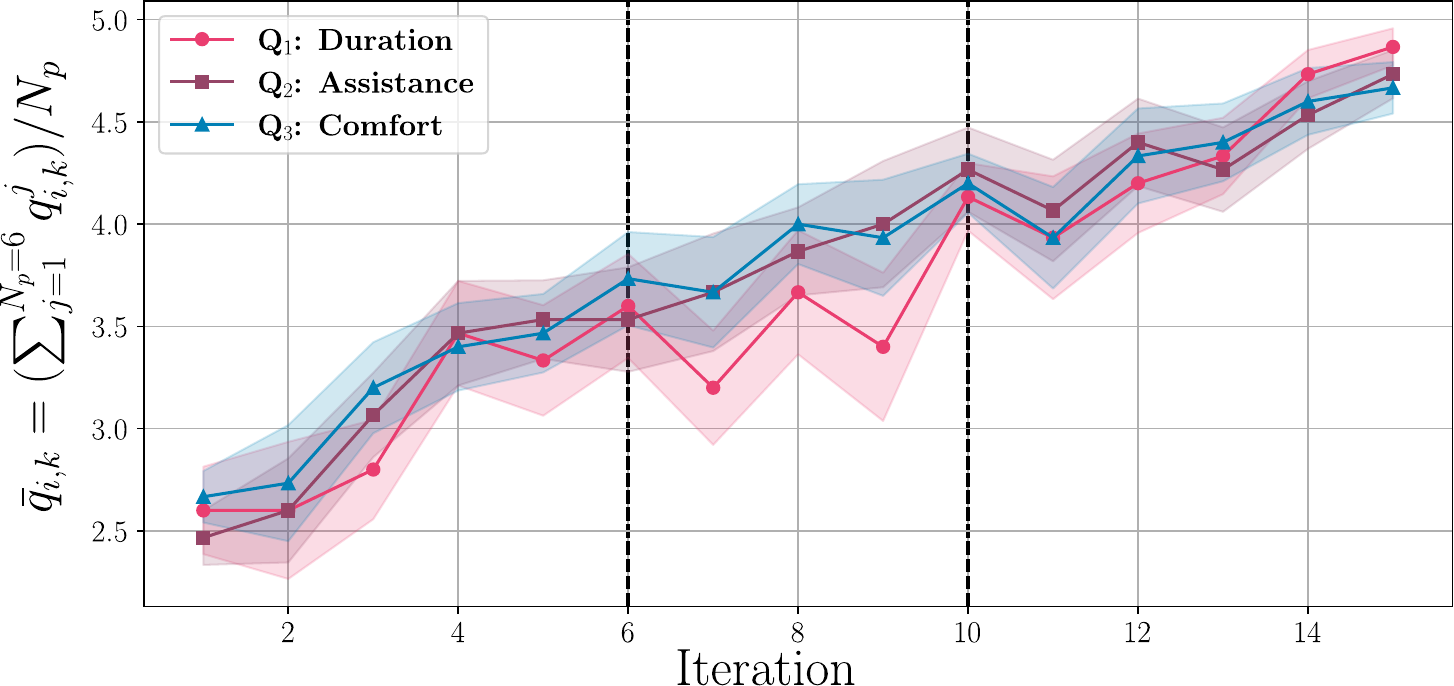}
\caption{Trend of the average scores $\bar{q}_{i,k}$. 
The \rev{central region exhibits} higher fluctuations, characteristic of the exploration of the search space. 
The dispersion bends represent the standard error of the mean.}
\label{fig:trends_questions}
\vspace{-0.2 cm}
\end{figure}

Figure \ref{fig:trends_questions} presents the results connected to the questions proposed to each participant, representing the average score $\bar{q}_{i,k}=(\sum_{j=1}^{N_p=6} q_{i,k}^j)/N_p$ for the $i$th question, given by the $j$th participant at the $k$th iteration.  
From the plot it is possible to notice the validity of the mechanism behind the iterative choice of the parameters: in the first iterations of the \emph{active learning} phase delimited by the dashed vertical lines, the trends present evident fluctuations, that are the direct consequence of testing unexplored regions of the search space, with the higher likelihood of proposing testing conditions that are far from the desired ones. 
In terms of subjective experience, the candidates reacted with preferences characterized by higher uncertainty.
However, all the three questions present an increasing trend, reaching values close to the maximum. This aspect, coupled with the scores for $A\%$ visible in Fig. \ref{fig:PBO_iterations}, \rev{allows us to conclude} that the framework is generally successful in optimizing the HRC painting application in a holistic sense.

\rev{\section{Conclusion}}
\label{sec:conclusions}

This work presented a novel human-centric collaborative framework that integrates Preference-Based Optimization (PBO) with Dynamic Movement Primitives (DMPs) to enhance \rev{HRC} in industrial painting applications. 
By redefining the robot's role from a primary executor to an adaptive assistant, our approach allows the system to adjust its behavior based on real-time input from the working environment, significantly reducing the required ranges of motion and enhancing the \rev{perceived} comfort. 
The proposed modifications to the DMPs framework 
proved effective in generating adaptive trajectories that respond to human preferences and \rev{ergonomics}.

\rev{Possible} future improvements are the inclusion of the points $\mathbf{p_A}$ and \rev{$\mathbf{p_C}$} into the set of design variables $\boldsymbol{\theta}$, in order to attenuate the problems connected to the initial mismatch of the operator's hand with respect to the piece. 
In addition, to account for slowdowns or fatigue of the operator, additional optimization and tracking modules may be added to correct the value of $\bar{\tau}$ based on modifications of the process conditions.

\printbibliography
\end{document}